\pdfoutput=1

\documentclass[11pt]{article}

\usepackage[]{acl}
\usepackage{adjustbox}
\usepackage{titlesec}
\usepackage{times}
\usepackage{latexsym}

\usepackage[T1]{fontenc}

\usepackage[utf8]{inputenc}
\usepackage{microtype}
\usepackage{array,multirow}
\title{Team QUST at SemEval-2023 Task 3: A Comprehensive Study of Monolingual and Multilingual Approaches for Detecting Online News Genre, Framing and Persuasion Techniques}

\author{Ye Jiang \\
  Qingdao University of Science and Technology, China \\
  \texttt{ye.jiang@qust.edu.cn} \\}

\begin{document}
\maketitle
\begin{abstract}
This paper describes the participation of team QUST in the SemEval2023 task 3. The monolingual models are first evaluated with the under-sampling of the majority classes in the early stage of the task. Then, the pre-trained multilingual model is fine-tuned with a combination of the class weights and the sample weights. Two different fine-tuning strategies, the task-agnostic and the task-dependent, are further investigated. All experiments are conducted under the 10-fold cross-validation, the multilingual approaches are superior to the monolingual ones. The submitted system\footnote{https://github.com/zgjiangtoby/SemEval2023\_QUST} achieves the second best in Italian and Spanish (zero-shot) in subtask-1. 
\end{abstract}

\section{Introduction}

Task 3 \citep{semeval2023task3} expects the participants to develop algorithms to automatically detect the news genre, framing and persuasion techniques in a multilingual setup as shown in Table \ref{examples}. Six different languages are covered in this task, including English, French, German, Italian, Polish and Russian. In addition, three surprise languages, Spanish, Greek and Georgian, are also included in the final test phase for conducting a zero-shot learning scenario. 

\begin{table*}
\centering
\begin{tabular}{l>{\raggedright}p{4.5cm}|l>{\raggedright}p{2.5cm}p{2cm}}
\hline
\textbf{Article-id (Para-id)} & \textbf{Text} & \multicolumn{3}{c}{\textbf{Labels}}\\
 & & \textbf{T1} & \textbf{T2} & \textbf{T3} \\
 \hline
710376094 (4) &  ...the asteroid named 3200 Phaethon ... is classified as “potentially hazardous” by the Minor Planet Center... & Opinion & Security and defense, 
Quality of life & Repetition\\
\hline
23114 (17) & Les études de l'Insee et de l'Ined confirment en outre que l'ascenseur social fonctionne...Chez les natifs (quatrième génération et plus)... & Reporting & Fairness and equality & Loaded Language  \\
\hline
2672 (3) & Ventisei milioni di cittadini chiusi in casa. Reclusi. Ai domiciliari. Gli abitanti di Shanghai... sono sottoposti a un lockdown severissimo & Satire & Policy prescription and evaluation, 
Quality of life & Loaded Language, 
Name Calling-Labeling \\
\hline
\end{tabular}
\caption{\label{examples}
Random examples from the dataset. T1 (subtask-1), T2 (subtask-2) and T3 (subtask-3) are document-level-multiclass, document-level-multilabel and sentence-level-multilabel tasks respectively, the paragraph id (Para-id) is only valid in subtask-3.}
\end{table*}

One of the major challenges of this task is that the model must have the ability to tackle the issues of imbalanced and insufficient training data. In subtask-1, the number of samples for labels `reporting', `opinion' and `satire' are 269, 878 and 87 respectively. With only few `satire' samples relative to other majority classes such as `opinion', the training model will easily overfit on majority classes and not learn enough from minority ones. For example, many batches will have no `satire' samples, so the gradients will be less informative during the training phase and result in poor performances in predicting the minority class. Meanwhile, previous studies \citep{leite2020toxic, zhangrevisiting} demonstrate that the scale of training data could influence the model performance significantly. However, the total training samples are 1234 in subtask-1, which is not enough to effectively train a deep learning model from scratch.

To address the above issues, traditional machine learning and deep learning methods are first evaluated with the under-sampling of the majority class in a monolingual setting. Then, the pre-trained multilingual model is fine-tuned to learn the cross-language features between languages. Specifically, fine-tuning is a common approach for utilizing the pre-trained language model (PLM) on many downstream tasks. It typically replaces the original output layer with a new task-specific layer and then fine-tunes the complete model. To model the features representation of news articles across different languages, the XLM-RoBERTa (XLM-R) \citep{conneau-etal-2020-unsupervised} is fine-tuned since it can processes all the languages existing in Task 3, and typically outperforms other models such as mBERT \citep{kenton2019bert}. 

In addition, this study also calculates the sample weights and class weights to combat the data imbalance. The sample weights enable a training batch that contains represents a good spread of the dataset. The class weights are able to penalize the misclassification made from the minority class by setting a higher class weight and reducing the weight for the majority class at the same time.

Two types of fine-tuning approaches, task-agnostic and task-dependent, are conducted on three subtasks. A task-agnostic strategy is a standard approach that fine-tunes the PLM on each subtask individually. This study also conducts a task-dependent strategy that fine-tunes the PLM from subtask-1 to subtask-3, since all the training samples have a similar content except the labels which are different. 

To evaluate, this paper conducts 10-fold cross-validation (CV) that comprehensively analyses the different monolingual and multilingual approaches and investigates the effectiveness of using sample weights and class weights to overcome the data imbalanced issue. The experimental results suggest that fine-tuning the PLM with sample weights and class weights can significantly improve the model performance in the task-dependent setting. In addition, the task-dependent fine-tuning strategy outperforms the task-agnostic method slightly, showing that the final fine-tuned model is able to learn the shared information between all the subtasks. 



\section{System Description}

The submitted system is a standard fine-tuning approach by adding a linear classifier to project the $[CLS]$ embedding to an unnormalized probability vector over the output classes. To address the imbalanced data issue, the class weights enable more weights to be assigned to the minority class in the cross-entropy loss. Therefore, the assigned weights are able to provide a higher penalty to the minority class and the algorithm could focus on reducing the errors for the minority class. 

Furthermore, the `satire' samples can be found with only 7\% of the training set. This might result in only few or even no minority class samples appearing in the training batch. To tackle this, the sample weights aim to calculate each sample's weights over the entire dataset, and then apply a weight to each sample to ensure that the majority class will have a smaller weight, or vice versa. The sample weights are also calculated in subtask-2 and subtask-3 except for the class weights. 

After training with all the positive strategies, early stopping is applied to save the best model in each fold. Eventually, the top 3 best models from the 10-fold CV are ensembled for the final prediction in each subtask.

\subsection{Weighted Fine-tuning}
The class weights are used to adjust the cross-entropy loss, which is the standard loss function to measure the difference between the predicted probability distribution and the true label distribution.

The class weights are a vector of values that correspond to each class label, and they are multiplied by the cross-entropy loss for each sample. Specifically, the class weights enable more weights to be assigned to the minority class, which means that the loss function will penalize the model more for misclassifying the minority class samples. Eventually, the model can learn to pay more attention to the minority class and improve its performance on the imbalanced data. Formally, given a total of $n$ training samples, the class weights $C_{w}^{j}$ for class $j$ can be simply calculated as:
\begin{equation}
    C_{w}^{j} = \frac{n}{c \times n_j}
\end{equation}
where $c$ denotes the number of classes, and $n_j$ is the total number of sample in class $j$. 

The class weight can be multiplied by the cross-entropy loss for each sample, to adjust the loss function. Specifically, let $y_c$ is the true label for class $j$, and $p_j$ is the predicted probability, then the cross-entropy loss for class $j$ is:
\begin{equation}
    L_j = -C_{w}^{j} \times y_j \times log(p_j)
\end{equation}
the total loss is the sum of the losses for all classes.

The weighting of classes is able to mitigate the impact of the data imbalance issue. However, given the size of the minority class is rather small, the training batches are likely to have only few or even no samples of the minority class and leading to retard of the model's convergence. 

To tackle this issue, the weighted random sampler is able to resample the training data, so that each batch contains a relatively balanced number of samples from each class. The weighted random sampler assigns a weight to each sample based on its class frequency, and then randomly selects samples from the dataset according to their weights.



\subsection{Task-Agnostic Vs. Task-Dependent}

The task-agnostic strategy is a standard approach that fine-tunes the PLM on each subtask individually. Specifically, the PLM is initialized with the pre-trained weights, and then trained on the specific data and labels of each subtask. The task-agnostic strategy can help the PLM to adapt to the domain and the task of each subtask, but it does not leverage the shared information or the hierarchical structure among the subtasks. 

The task-dependent strategy is an alternative approach that fine-tunes the PLM from subtask-1 to subtask-3, since all the training samples have the same content except the labels are different. This means that the PLM is initialized with the pre-trained weights, and then trained on the data and labels of subtask-1. After that, the PLM is further trained on the data and labels of subtask-2, using the weights from subtask-1 as the initial weights. Similarly, the PLM is further trained on the data and labels of subtask-3, using the weights from subtask-2 as the initial weights. The task-dependent strategy can help the PLM to transfer the knowledge and the features learned from the previous subtasks to the next subtasks.

\section{Experimental Setup}

\subsection{Data}

The total number of news articles in subtask-1 and subtask-2 are 1234 and 1238 respectively. The subtask-3 divides each article into paragraphs, so the total number of paragraphs is 20207. The maximum, minimum and average numbers of tokens in three subtasks are calculated as shown in Table \ref{statistic}. All the experiments discussed in this paper were conducted with 10-fold CV, and the results displayed are the averages. Specifically, the dataset is split into 10 folds for each subtask, and the same folds are used for the task-agnostic strategy and the task-dependent strategy. For each fold, one fold is used as the test set and the other nine folds are used as the training set. 

\begin{table}
\centering
\begin{adjustbox}{max width=0.48\textwidth}
\begin{tabular}{lccc}
\hline
  & T1 & T2 & T3\\
\hline
minimum & 88 & 67 & 3  \\
maximum & 7747 & 7747 & 1069 \\
average & 763 & 761 & 47 \\
\hline
\end{tabular}
\end{adjustbox}
\caption{Minimum, maximum and average numbers of tokens in each subtask.}
\label{statistic}
\end{table}

\subsection{Preprocessing}
The punctuation, links, escape characters (e.g., `\textbackslash n'), and numbers are removed from the data, and strings are lowercased. All news articles are truncated or padded to a maximum of 512 tokens and different languages are merged into a JSON file. In subtask-3, a `None' class is added to represent the paragraphs that do not have a label. 

In data version three (v3), the organizer released the labels of the development set. Intuitively, it would be beneficial to the model performance by expanding the scale of training data. Therefore, the v3 merges the training and development data from all the languages for conducting a 10-fold CV. The results of subtask-1 are evaluated with macro-f1, and the other subtasks are evaluated with micro-f1.

\subsection{Model Configurations}

Several baseline models are compared before the final system is built. Given the scale of the training data is relatively small, traditional machine learning approaches are first evaluated in a monolingual setting. All traditional approaches are built based on TF-IDF vectors with under-sampling the majority class. Then, deep learning methods include CNN \citep{jiang2019team}, RNN \citep{linstructured} and Self-attention \citep{jiang2022topic} are built upon with the $[CLS]$ embedding from the monolingual PLMs respectively. Both traditional and deep learning methods are conducted with \textit{imblearn} \citep{JMLR:v18:16-365} and \textit{skorch} \cite{skorch}. Finally, the task-agnostic and the task-dependent of the fine-tuned PLMs are also evaluated.  

The final system is built upon the pre-trained XLM-RoBETRa by using the huggingface library. The AdamW \cite{loshchilov2017decoupled} optimizer is utilized with a learning rate of 3e-5, and training epochs are 30. In each fold, the early stopping tolerance is 5 epochs, and only the model with the highest score is saved. The top three best systems are saved in each subtask, and the final predictions are made by ensembling\footnote{The systems are ensembled through the majority voting.} all the predictions from the top three models. 

\section{Results}

\subsection{Monolingual Approaches}
The evaluations are conducted in two stages (i.e., the v2 and v3 of the released data). In the v2 stage, the evaluations aim to find out if the monolingual approaches are effective given the scale of training data is relatively small in each language. The monolingual models are first evaluated with traditional machine learning approaches and deep learning approaches as shown in Table \ref{result1}. The v2 stage only evaluates subtask-1 in order to have a quick `warmup' and find out the feasible approaches between models and data. 

\begin{table}
\centering
\begin{adjustbox}{max width=0.48\textwidth}
\begin{tabular}{l|llllll}
\hline
 \textbf{Models} & \textbf{en} & \textbf{fr} & \textbf{de} & \textbf{it} & \textbf{po} & \textbf{ru} \\
\hline
SVM & 0.36 & 0.37 & 0.31 & 0.16 & 0.38 & 0.14  \\
LG & 0.42 & \textbf{0.48} & 0.42 & \textbf{0.29} & \textbf{0.67} & \textbf{0.29} \\
RF & \textbf{0.49} & 0.32 & 0.33 & 0.27 & 0.50 & 0.24 \\
\hline
BERT-CNN & 0.21 & 0.37 & 0.37 & 0.25 & 0.47 & 0.23 \\
BERT-RNN & 0.19 & 0.35 & 0.35 & 0.19 & 0.49 & 0.21 \\
BERT-Self & 0.26 & 0.31 & 0.28 & 0.13 & 0.37 & 0.22 \\
RBERT-CNN & 0.41 & 0.43 & \textbf{0.44}  & \textbf{0.29} & 0.58 & 0.25\\
RBERT-RNN & 0.29 & 0.35 & \textbf{0.44} & 0.23 & 0.55 & 0.21 \\
RBERT-Self & 0.24 & 0.35 & 0.41 & 0.23 & 0.55 & 0.17\\

\hline
\end{tabular}
\end{adjustbox}
\caption{Macro-f1 of the monolingual models on the v2 subtask-1. SVM, LG and RF denote Support Vector Machine, Logistic Regression and Random Forest respectively. The CNN, RNN and Self models are built based on the $[CLS]$ embeddings directly encoded from BERT and RBERT (RoBERTa). }
\label{result1}
\end{table}

As a result, traditional machine learning approaches outperform deep learning approaches in general. For example, a simple combination of TF-IDF and a logistic regression classifier with an under-sample of the majority class achieves the best score in four out of six languages (i.e., fr, it, po and ru). Meanwhile, a similar combination but with a random forest classifier yields the best score on the dev leaderboard during the v2 stage. 

In terms of deep learning methods, all models are first using the PLMs\footnote{The complete list of PLMs in monolingual settings are shown in Appendix \ref{appendix_a}.} to encode the news article and obtain the $[CLS]$ embeddings. Then, the $[CLS]$ embeddings are taken as the input to the multi-kernel CNN, attentive RNN and self-attention networks. On the one hand, the overall performance of deep learning approaches is sub-optimal to the traditional ones. One possible reason is that the deep learning models with a comparatively large number of parameters in relation to the small size of the training set result in overfitting. On the other hand, the use of RoBERTa can yield better performance than BERT in subtask-2. In addition, the CNN models are generally better than attentive RNN and self-attention in all sub-tasks. 

\subsection{Multilingual Approaches}
Although the monolingual approaches achieve good performance in a single language, the performances are not consistent in all languages. In order to capture the shared information between all languages and train a multilingual model which can process all the languages at once, the pre-trained XLM-R is utilized in the v3 stage as shown in Table \ref{result2}.

\begin{table}
\centering
\begin{adjustbox}{max width=0.48\textwidth}
\begin{tabular}{l|ccc|ccc}
\hline
 \textbf{Languages} & \multicolumn{3}{c}{\textbf{XLM-R(v2)}} & \multicolumn{3}{c}{\textbf{XLM-R(v3)}} \\
 & T1 & T2 & T3 & T1 & T2 & T3 \\
 \hline
en & 0.51 & 0.50 & 0.47 & 0.51 & 0.52 & 0.49 \\
fr & 0.62 & 0.41 & 0.30 & 0.62 & 0.42 & 0.29 \\
de & 0.64 & 0.63 & 0.34 & 0.65 & 0.65 & 0.37 \\
it & 0.75 & 0.54 & 0.30 & 0.78 & 0.54 & 0.31 \\
po & 0.50 & 0.59 & 0.15 & 0.51 & 0.59 & 0.13 \\
ru & 0.47 & 0.30 & 0.20 & 0.49 & 0.27 & 0.21 \\
\hline
\end{tabular}
\end{adjustbox}
\caption{Average scores (i.e., macro-f1 for T1 and micro-f1 for T2 and T3) of XLM-RoBERTa (XLM-R) from 10-fold CV on each task based on task-dependent fine-tuning strategy. }
\label{result2}
\end{table}

The multilingual approaches are evaluated in both the v2 and v3 stages. In the v2 stage, the XLM-R is fine-tuned based on a 10-fold CV from the released v2 dataset. To combat the imbalanced data during fine-tuning, the sample weights and the class weights are calculated. The weighted random sampler is combined with sample weights to balance the training batches relatively. The class weights are multiplied by the training loss to make the model concern more on the minority class. 

As a result, the performance is significantly improved compared with the monolingual approaches in the v2 stage (see XLM-R(v2) T1 column), showing that some languages (e.g., it, fr and en) can capture meaningful features from other languages. Given the news articles have the same content but the labels are different between subtasks, the best model obtained from subtask-1 is further fine-tuned on subtask-2 and subtask-3. 

In the v3 stage, the released labels from the development dataset enable an expansion of the training samples in the 10-fold CV. Therefore, the same fine-tuning method is retained except the training samples become larger in the v3 stage. The results from XLM-R(v3) indicate that the expansion of the training sample leads to a slight performance improvement. Finally, the top 3 best XLM-R models from the v3 are ensembled to make final predictions.

\subsection{Ablation Study}
In order to investigate the effectiveness of each component in the proposed method, an ablation study is conducted as shown in Table \ref{result3}. The results indicate that the use of the class weights, the sample weights and the task-dependent fine-tuning can significantly improve the model performance in all subtasks.

Specifically, the average macro-f1 is dropped by 14\% when the class weights are removed. A quick manual inspection of the predicted files also finds that most of the news articles are predicted as `opinion', which is the largest majority class of the training data. The scores get worse when the sample weights are removed as well. This might be because some of the training batches do not contain any minority class sample, and the less informative features are updated during the training phase. Then, the task-dependent is removed and the XLM-R(v3) w/o td becomes a vanila XLM-R model that does not utilize any of the strategies from the above. This leads to the fine-tuning strategy becoming task-agnostic. The results demonstrate that the task-dependent fine-tuning (i.e., XLM-R(v3) w/o sw) learns the shared information between tasks and leads to better performance than the task-agnostic one (i.e., XLM-R(v3) w/o td).

\begin{table}
\centering
\begin{adjustbox}{max width=0.48\textwidth}
\begin{tabular}{l|ccc}
\hline
 \textbf{Methods} & T1 & T2 & T3  \\
 \hline
XLM-R(v3)  & 0.51 & 0.52 & 0.49  \\
XLM-R(v3) w/o cw & 0.37 & n/a & n/a  \\
XLM-R(v3) w/o sw  & 0.25 & 0.31 & 0.20  \\
XLM-R(v3) w/o td  & 0.24 & 0.31 & 0.19  \\
\hline
\end{tabular}
\end{adjustbox}
\caption{Ablation study of the fine-tuned XLM-R model. cw, sw and td denote class weights, sample weights and task-dependent respectively. Note that the class weights are not used in subtask-2 and subtask-3.}
\label{result3}
\end{table}

\subsection{Error Analysis}
\begin{figure}[h]
\includegraphics[width=0.48\textwidth]{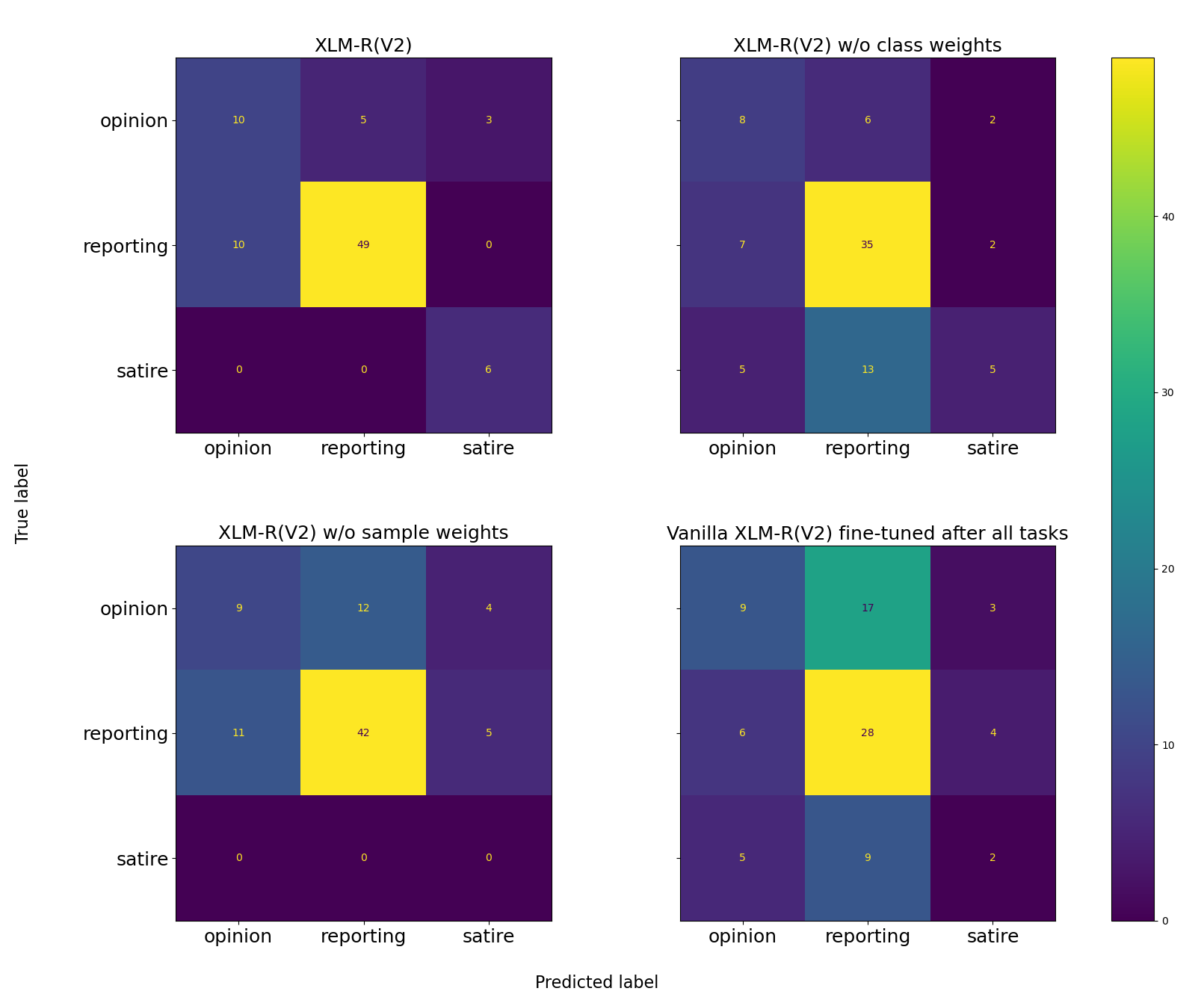}
\caption{Confusion matrix of each component of the XLM-R(v2) in English subtask-1.}
\label{fig1}
\end{figure}

In the error analysis, the XLM-R(v2) is fine-tuned only on the training set of v2 data with 10-fold CV, and uses the best model to predict the labels of the v3 development data. The confusion matrix of each component in the XLM-R(v2) is shown in Figure \ref{fig1}.

The XLM-R(v2) achieves the highest number of correct predictions on the minority class `satire' as well as the majority class `reporting'. However, the performance decayed significantly when the class weights and the sample weights are removed. The sample weights removed XLM-R(v2) can not correctly predict the minority class `satire', showing that the standard fine-tuning is not able to make correct predictions when the data is highly imbalanced. In order to evaluate the effectiveness of the shared information between subtasks, the last XLM-R(v2) removes the class weights and the sample weights, and is directly fine-tuned through three subtasks. As a result, the last XLM-R(v2) obtains the least number of correct predictions on `reporting', and makes the highest number of misclassifications between `reporting' and `opinion'. However, it obtains two correct predictions on the minority class `satire', showing that the shared information is able to provide useful features to handle the minority class.

\subsection{Results on the official test set}

Team QUST ranked 2nd in Italian and Spanish (zero-shot) on the T1 official test set as shown in Table \ref{result_all}. In addition, the proposed method surpasses T1 and T2 baselines in all languages except that in German and in Polish. Interestingly, the proposed method performs significantly bad in T3, and all submissions fail to the T3 baselines. The possible reasons could be two-fold: 1) All early-stage evaluations are conducted based on model performances from T1, therefore the results can not fully reflect that of T3. 2) T1 and T2 are both document-level classification tasks, but T3 is sentence-level so the T2 could benefit more than T3 from the task-dependent fine-tuning.

\begin{table}
\centering
\begin{adjustbox}{max width=0.48\textwidth}
\begin{tabular}{l|ccc|ccc}
\hline
 \textbf{Lan} & \multicolumn{3}{c}{\textbf{Baseline}} & \multicolumn{3}{c}{\textbf{QUST} (rank)} \\
 \hline
 & T1 & T2 & T3 & T1 & T2 & T3 \\
 \hline
en & 0.29 & 0.35 & \textbf{0.20} & \textbf{0.51} (10th) & \textbf{0.51} (6th) & 0.14 (22nd) \\
fr & 0.57 & 0.33 & \textbf{0.24} & \textbf{0.62} (9th) & \textbf{0.45} (11th) & 0.21 (19th) \\
de &  \textbf{0.63} & 0.49 & \textbf{0.32} & \textbf{0.63} (10th) & \textbf{0.62} (7th) & 0.15 (19th) \\
it & 0.39 & 0.49 & \textbf{0.40} & \textbf{0.77} (2nd) & \textbf{0.50} (10th) & 0.21 (19th) \\
po & 0.49 & \textbf{0.59} & \textbf{0.18} & \textbf{0.53} (11th) & \textbf{0.59} (11th) & 0.10 (19th) \\
ru & 0.40 & 0.23 & \textbf{0.21} & \textbf{0.47} (9th) & \textbf{0.25} (12th) & 0.10 (19th) \\
es & 0.15 & 0.12 & \textbf{0.25} & \textbf{0.55} (2nd) & \textbf{0.37} (13th) & 0.12 (17th) \\
gr & 0.17 & 0.35 & \textbf{0.08} & \textbf{0.49} (10th) & \textbf{0.41} (9th) & 0.05 (15th) \\
ka & 0.26 & 0.26 & \textbf{0.14} & \textbf{0.54} (7th) & \textbf{0.31} (12th) & 0.09 (15th) \\
\hline
\end{tabular}
\end{adjustbox}
\caption{Official Task 3 leaderboard on the test set. Better scores are bold compared with baselines.}
\label{result_all}
\end{table}
 
\section{Conclusion}
This paper extensively describes the monolingual and multilingual approaches in detecting news genre, news framing and persuasion techniques. To address the issue of imbalanced data, the majority classes are first under-sampled at the early stage of the task. The traditional machine learning models and deep learning models are evaluated in a monolingual setup. In order to learn the shared information between all languages, a multilingual model is built in the later stage of the task. The XLM-R is fine-tuned with the pre-calculated class weights and sample weights to combat the imbalanced data. In addition, two types of fine-tuning strategies, the task-agnostic and the task-dependent, are evaluated respectively. The proposed method is the second best system in Italian and in Spanish (zero-shot) subtask-1. In future work, the minority class sample can be back-translated from other languages to augment the data size. 






\bibliography{anthology,custom}

\begin{thebibliography}{16}
\expandafter\ifx\csname natexlab\endcsname\relax\def\natexlab#1{#1}\fi

\bibitem[{Conneau et~al.(2020)Conneau, Khandelwal, Goyal, Chaudhary, Wenzek,
  Guzm{\'a}n, Grave, Ott, Zettlemoyer, and
  Stoyanov}]{conneau-etal-2020-unsupervised}
Alexis Conneau, Kartikay Khandelwal, Naman Goyal, Vishrav Chaudhary, Guillaume
  Wenzek, Francisco Guzm{\'a}n, Edouard Grave, Myle Ott, Luke Zettlemoyer, and
  Veselin Stoyanov. 2020.
\newblock \href {https://doi.org/10.18653/v1/2020.acl-main.747} {Unsupervised
  cross-lingual representation learning at scale}.
\newblock In \emph{Proceedings of the 58th Annual Meeting of the Association
  for Computational Linguistics}, pages 8440--8451, Online. Association for
  Computational Linguistics.

\bibitem[{Devlin et~al.(2018)Devlin, Chang, Lee, and
  Toutanova}]{devlin2018bert}
Jacob Devlin, Ming-Wei Chang, Kenton Lee, and Kristina Toutanova. 2018.
\newblock Bert: Pre-training of deep bidirectional transformers for language
  understanding.
\newblock \emph{arXiv preprint arXiv:1810.04805}.

\bibitem[{Jiang et~al.(2019)Jiang, Petrak, Song, Bontcheva, and
  Maynard}]{jiang2019team}
Ye~Jiang, Johann Petrak, Xingyi Song, Kalina Bontcheva, and Diana Maynard.
  2019.
\newblock Team bertha von suttner at semeval-2019 task 4: Hyperpartisan news
  detection using elmo sentence representation convolutional network.
\newblock In \emph{Proceedings of the 13th International Workshop on Semantic
  Evaluation}, pages 840--844.

\bibitem[{Jiang and Wang(2022)}]{jiang2022topic}
Ye~Jiang and Yimin Wang. 2022.
\newblock Topic-aware hierarchical multi-attention network for text
  classification.
\newblock \emph{International Journal of Machine Learning and Cybernetics},
  pages 1--13.

\bibitem[{Kenton and Toutanova(2019)}]{kenton2019bert}
Jacob Devlin Ming-Wei~Chang Kenton and Lee~Kristina Toutanova. 2019.
\newblock Bert: Pre-training of deep bidirectional transformers for language
  understanding.
\newblock In \emph{Proceedings of NAACL-HLT}, pages 4171--4186.

\bibitem[{Kuratov and Arkhipov(2019)}]{kuratov2019adaptation}
Yuri Kuratov and Mikhail Arkhipov. 2019.
\newblock Adaptation of deep bidirectional multilingual transformers for
  russian language.
\newblock \emph{arXiv preprint arXiv:1905.07213}.

\bibitem[{Leite et~al.(2020)Leite, Silva, Bontcheva, and
  Scarton}]{leite2020toxic}
Joao~A Leite, Diego~F Silva, Kalina Bontcheva, and Carolina Scarton. 2020.
\newblock Toxic language detection in social media for brazilian portuguese:
  New dataset and multilingual analysis.
\newblock \emph{arXiv preprint arXiv:2010.04543}.

\bibitem[{Lema{{\^i}}tre et~al.(2017)Lema{{\^i}}tre, Nogueira, and
  Aridas}]{JMLR:v18:16-365}
Guillaume Lema{{\^i}}tre, Fernando Nogueira, and Christos~K. Aridas. 2017.
\newblock \href {http://jmlr.org/papers/v18/16-365} {Imbalanced-learn: A python
  toolbox to tackle the curse of imbalanced datasets in machine learning}.
\newblock \emph{Journal of Machine Learning Research}, 18(17):1--5.

\bibitem[{Lin et~al.(2017)Lin, Feng, dos Santos, Yu, Xiang, Zhou, and
  Bengio}]{linstructured}
Zhouhan Lin, Minwei Feng, Cicero~Nogueira dos Santos, Mo~Yu, Bing Xiang, Bowen
  Zhou, and Yoshua Bengio. 2017.
\newblock A structured self-attentive sentence embedding.
\newblock In \emph{International Conference on Learning Representations}.

\bibitem[{Liu et~al.(2019)Liu, Ott, Goyal, Du, Joshi, Chen, Levy, Lewis,
  Zettlemoyer, and Stoyanov}]{liu2019roberta}
Yinhan Liu, Myle Ott, Naman Goyal, Jingfei Du, Mandar Joshi, Danqi Chen, Omer
  Levy, Mike Lewis, Luke Zettlemoyer, and Veselin Stoyanov. 2019.
\newblock Roberta: A robustly optimized bert pretraining approach.
\newblock \emph{arXiv preprint arXiv:1907.11692}.

\bibitem[{Loshchilov and Hutter(2017)}]{loshchilov2017decoupled}
Ilya Loshchilov and Frank Hutter. 2017.
\newblock Decoupled weight decay regularization.
\newblock \emph{arXiv preprint arXiv:1711.05101}.

\bibitem[{Minixhofer et~al.(2022)Minixhofer, Paischer, and
  Rekabsaz}]{minixhofer-etal-2022-wechsel}
Benjamin Minixhofer, Fabian Paischer, and Navid Rekabsaz. 2022.
\newblock \href {https://doi.org/10.18653/v1/2022.naacl-main.293} {{WECHSEL}:
  Effective initialization of subword embeddings for cross-lingual transfer of
  monolingual language models}.
\newblock In \emph{Proceedings of the 2022 Conference of the North American
  Chapter of the Association for Computational Linguistics: Human Language
  Technologies}, pages 3992--4006, Seattle, United States. Association for
  Computational Linguistics.

\bibitem[{Piskorski et~al.(2023)Piskorski, Stefanovitch, Da~San~Martino, and
  Nakov}]{semeval2023task3}
Jakub Piskorski, Nicolas Stefanovitch, Giovanni Da~San~Martino, and Preslav
  Nakov. 2023.
\newblock Semeval-2023 task 3: Detecting the category, the framing, and the
  persuasion techniques in online news in a multi-lingual setup.
\newblock In \emph{Proceedings of the 17th International Workshop on Semantic
  Evaluation}, SemEval 2023, Toronto, Canada.

\bibitem[{Schweter(2020)}]{stefan_schweter_2020_4275044}
Stefan Schweter. 2020.
\newblock \href {https://doi.org/10.5281/zenodo.4275044} {Europeana bert and
  electra models}.

\bibitem[{Tietz et~al.(2017)Tietz, Fan, Nouri, Bossan, and {skorch
  Developers}}]{skorch}
Marian Tietz, Thomas~J. Fan, Daniel Nouri, Benjamin Bossan, and {skorch
  Developers}. 2017.
\newblock \href {https://skorch.readthedocs.io/en/stable/} {\emph{skorch: A
  scikit-learn compatible neural network library that wraps PyTorch}}.

\bibitem[{Zhang et~al.(2021)Zhang, Wu, Katiyar, Weinberger, and
  Artzi}]{zhangrevisiting}
Tianyi Zhang, Felix Wu, Arzoo Katiyar, Kilian~Q Weinberger, and Yoav Artzi.
  2021.
\newblock Revisiting few-sample bert fine-tuning.
\newblock In \emph{International Conference on Learning Representations}.

\end{thebibliography}
\bibliographystyle{acl_natbib}

\appendix
\section{PLMs in monolingual settings}\label{appendix_a}
The Huggingface model names used in monolingual experiments are shown as follows:
\begin{table}[h!]
    \centering
    \begin{adjustbox}{max width=0.48\textwidth}
    \begin{tabular}{c|c}
    \hline
    Lan & BERT model names \\
    \hline
       en  &  bert-base-cased\cite{devlin2018bert} \\
       fr  &  dbmdz/bert-base-french-europeana-cased\cite{stefan_schweter_2020_4275044}  \\
       de  &  bert-base-german-cased   \\
       it  &  dbmdz/bert-base-italian-cased\cite{stefan_schweter_2020_4275044}    \\
       po  &  dkleczek/bert-base-polish-uncased-v1    \\
       ru  &  DeepPavlov/rubert-base-cased\cite{kuratov2019adaptation}  \\
    \hline
    \end{tabular}
    \end{adjustbox}

\end{table}

\begin{table}[h!]
    \centering
    \begin{adjustbox}{max width=0.48\textwidth}
    \begin{tabular}{c|c}
    \hline
    Lan & RoBERTa model names \\
    \hline
       en  & roberta-base\cite{liu2019roberta} \\
       fr  & benjamin/roberta-base-wechsel-french\cite{minixhofer-etal-2022-wechsel} \\
       de  & benjamin/roberta-base-wechsel-german\cite{minixhofer-etal-2022-wechsel}  \\
       it  & flax-community/robit-roberta-base-it   \\
       po  & clarin-pl/roberta-polish-kgr10   \\
       ru  & blinoff/roberta-base-russian-v0 \\
    \hline
    \end{tabular}
    \end{adjustbox}

\end{table}

\end{document}